\documentclass[10pt,twocolumn,letterpaper]{article}

\usepackage[pagenumbers]{cvpr} 

\usepackage{graphicx}
\usepackage{amsmath}
\usepackage{amssymb}
\usepackage{booktabs}
\usepackage{multirow}
\usepackage{tabularx}
\usepackage{float}
\usepackage{xcolor}
\usepackage{bbm}
\usepackage{import}
\usepackage{mathrsfs}

\makeatletter
\@namedef{ver@everyshi.sty}{}
\makeatother
\usepackage{tikz}

\usepackage{comment}

\usepackage{pifont} 
\usepackage{booktabs}
\usepackage{multirow}
\usepackage{adjustbox}
\usepackage{fontawesome} 
\usepackage{amsmath} 

\definecolor{opA}{rgb}{0.9,0.6,0.0}
\definecolor{opB}{rgb}{0.35,0.70,0.90}
\definecolor{opC}{rgb}{0.8,0.40,0.0}
\definecolor{opD}{rgb}{0.0,0.60,0.50} %
\definecolor{opE}{rgb}{0.8,0.6,0.7}
\definecolor{opF}{rgb}{0.,0.45,0.70} 

\definecolor{pltBlue}{rgb}{0.12156862745098039, 0.4666666666666667, 0.7058823529411765}
\definecolor{pltOrange}{rgb}{1.0, 0.4980392156862745, 0.054901960784313725}
\definecolor{pltGreen}{rgb}{0.17254901960784313, 0.6274509803921569, 0.17254901960784313}
\definecolor{pltRed}{rgb}{0.8392156862745098, 0.15294117647058825, 0.1568627450980392}
\definecolor{pltViolet}{rgb}{0.5803921568627451, 0.403921568627451, 0.7411764705882353}
\definecolor{pltBrown}{rgb}{0.5490196078431373, 0.33725490196078434, 0.29411764705882354}
\definecolor{pltMagenta}{rgb}{0.8901960784313725, 0.4666666666666667, 0.7607843137254902}
\definecolor{pltGray}{rgb}{0.4980392156862745, 0.4980392156862745, 0.4980392156862745}
\definecolor{pltLightGreen}{rgb}{0.7372549019607844, 0.7411764705882353, 0.13333333333333333}
\definecolor{pltCyan}{rgb}{0.09019607843137255, 0.7450980392156863, 0.8117647058823529}
\definecolor{pltPink}{rgb}{0.49803921568627, 0.49803921568627, 0.49803921568627}

\def\link#1{
    \ifx&#1&
        \xmark{}
    \else
        {\href{#1}{\faExternalLink}}
    \fi
}

\newcommand{\methodtitle}{Imitator: Personalized Speech-driven 3D Facial Animation\xspace}

\usepackage[pagebackref,breaklinks,colorlinks]{hyperref}

\usepackage[capitalize]{cleveref}
\crefname{section}{Sec.}{Secs.}
\Crefname{section}{Section}{Sections}
\Crefname{table}{Table}{Tables}
\crefname{table}{Tab.}{Tabs.}

\begin{document}
\title{\methodtitle}

\author{
Balamurugan Thambiraja\textsuperscript{1}
\qquad 
Ikhsanul Habibie\textsuperscript{2}
\qquad 
Sadegh Aliakbarian\textsuperscript{3}
\\
Darren Cosker\textsuperscript{3}
\qquad
Christian Theobalt\textsuperscript{2}
\quad
Justus Thies\textsuperscript{1}
\\
\\
\textsuperscript{1} Max Planck Institute for Intelligent Systems, Tübingen, Germany \\
\textsuperscript{2} Max Planck Institute for Informatics, Saarland, Germany \\
\textsuperscript{3} Microsoft Mixed Reality \& AI Lab, Cambridge, UK
}

\twocolumn[{%
\renewcommand\twocolumn[1][]{#1}%
\maketitle
\begin{center}
    \centering
    \captionsetup{type=figure}
    \vspace{-0.5cm}
    \includegraphics[width=\textwidth]{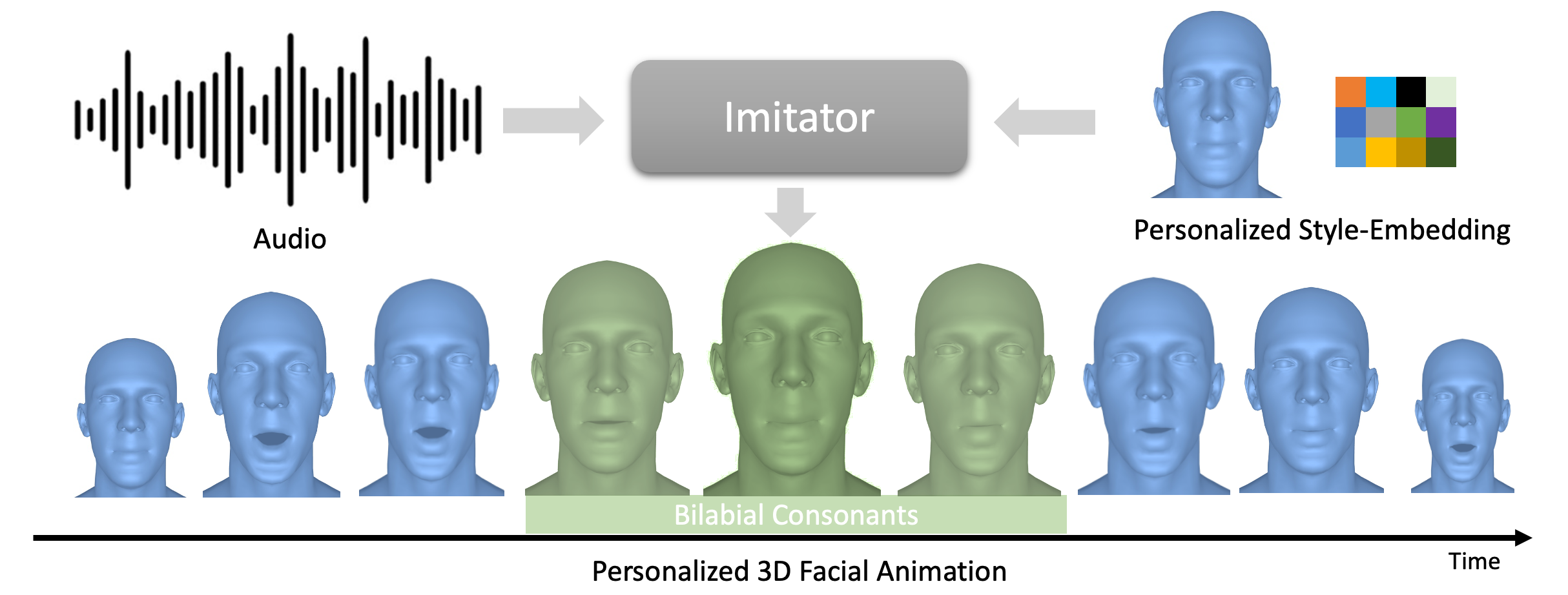}
      \caption{
      \textit{Imitator} is a novel method for personalized speech-driven 3D facial animation.
      Given an audio sequence and a personalized style-embedding as input, we generate person-specific motion sequences with accurate lip closures for bilabial consonants ('m','b','p').
      The style-embedding of a subject can be computed by a short reference video (e.g., 5s).
      }
      \label{fig:teaser}
\end{center}%
}]

\renewcommand{\paragraph}[1]{\vspace{0.1cm}\noindent\textbf{#1}}

\begin{abstract}
Speech-driven 3D facial animation has been widely explored, with applications in gaming, character animation, virtual reality, and telepresence systems.
State-of-the-art methods deform the face topology of the target actor to sync the input audio without considering the identity-specific speaking style and facial idiosyncrasies of the target actor, thus, resulting in unrealistic and inaccurate lip movements.
To address this, we present Imitator, a speech-driven facial expression synthesis method, which learns identity-specific details from a short input video and produces novel facial expressions matching the identity-specific speaking style and facial idiosyncrasies of the target actor.
Specifically, we train a style-agnostic transformer on a large facial expression dataset which we use as a prior for audio-driven facial expressions.
Based on this prior, we optimize for identity-specific speaking style based on a short reference video.
To train the prior, we introduce a novel loss function based on detected bilabial consonants to ensure plausible lip closures and consequently improve the realism of the generated expressions.
Through detailed experiments and a user study, we show that our approach produces temporally coherent facial expressions from input audio while preserving the speaking style of the target actors.
Please check out the project \href{https://balamuruganthambiraja.github.io/Imitator}{page} for the supplemental video and more results. 
\end{abstract}

\section{Introduction}
\label{sec:intro}
3D digital humans raised a lot of attention in the past few years as they aim to replicate the appearance and motion of real humans for immersive applications, like telepresence in AR or VR, character animation and creation for entertainment (movies and games), and virtual mirrors for e-commerce.
Especially, with the introduction of neural rendering~\cite{tewari2022advances,tewari2020neuralrendering}, we see immense progress in the photo-realistic synthesis of such digital doubles~\cite{imavatar, nerface, neural-volumes}.
These avatars can be controlled via visual tracking to mirror the facial expressions of a real human.
However, we need to control the facial avatars with text or audio inputs for a series of applications.
For example, AI-driven digital assistants rely on motion synthesis instead of motion cloning.
Even telepresence applications might need to work with audio inputs only, when the face of the person is occluded or cannot be tracked, since a face capture device is not available.
To this end, we analyze motion synthesis for facial animations from audio inputs; note that text-to-speech approaches can be used to generate such audio.
Humans are generally sensitive towards faces, especially facial motions, as they are crucial for communication (e.g., micro-expressions).
Without full expressiveness and proper lip closures, the generated animation will be perceived as unnatural and implausible.
Especially if the person is known, the facial animations must match the subject's idiosyncrasies.
Recent methods for speech-driven 3D facial animation~\cite{karras_audio-driven_2017, voca, meshtalk, faceformer} are data-driven.
They are trained on high-quality motion capture data and leverage pretrained speech models~\cite{deepspeech, wav2vec} to extract an intermediate audio representation.
We can classify these data-driven methods into two categories, generalized~\cite{voca, meshtalk, faceformer} and personalized animation generation methods~\cite{karras_audio-driven_2017}.
In contrast to those approaches, we aim at a personalized 3D facial animation synthesis that can adapt to a new user while only relying on input RGB videos captured with commodity cameras.
Specifically, we propose a transformer-based auto-regressive motion synthesis method that predicts a generalized motion representation.
This intermediate representation is decoded by a motion decoder which is adaptable to new users.
A speaker embedding is adjusted for a new user, and a new motion basis for the motion decoder is computed.
Our method is trained on the VOCA dataset~\cite{voca} and can be applied to new subjects captured in a short monocular RGB video.
As lip closures are of paramount importance for bilabial consonants ('m','b','p'), we introduce a novel loss based on the detection of bilabials to ensure that the lips are closed properly.
We take inspiration from the locomotion synthesis field~\cite{lee2002interactive, holden2016deep}, where similar losses are used to enforce foot contact with the ground and transfer it to our scenario of physically plausible lip motions.

In a series of experiments and ablation studies, we demonstrate that our method is able to synthesize facial expressions that match the target subject's motions in terms of style and expressiveness.
Our method outperforms state-of-the-art methods in our metrical evaluation and user study. Please refer to our supplemental video for a detailed qualitative comparison.
In a user study, we confirm that personalized facial expressions are important for the perceived realism.

\medskip
\noindent
The contributions of our work \textit{Imitator} are as follows:
\begin{itemize}
    \item a novel auto-regressive motion synthesis architecture that allows for adaption to new users by disentangling generalized viseme generation and person-specific motion decoding,
    \item and a lip contact loss formulation for improved lip closures based on physiological cues of bilabial consonants ('m','b','p').
\end{itemize}
\begin{figure*}[ht!]
    \vspace{-0.25cm}
    \includegraphics[width=\textwidth]{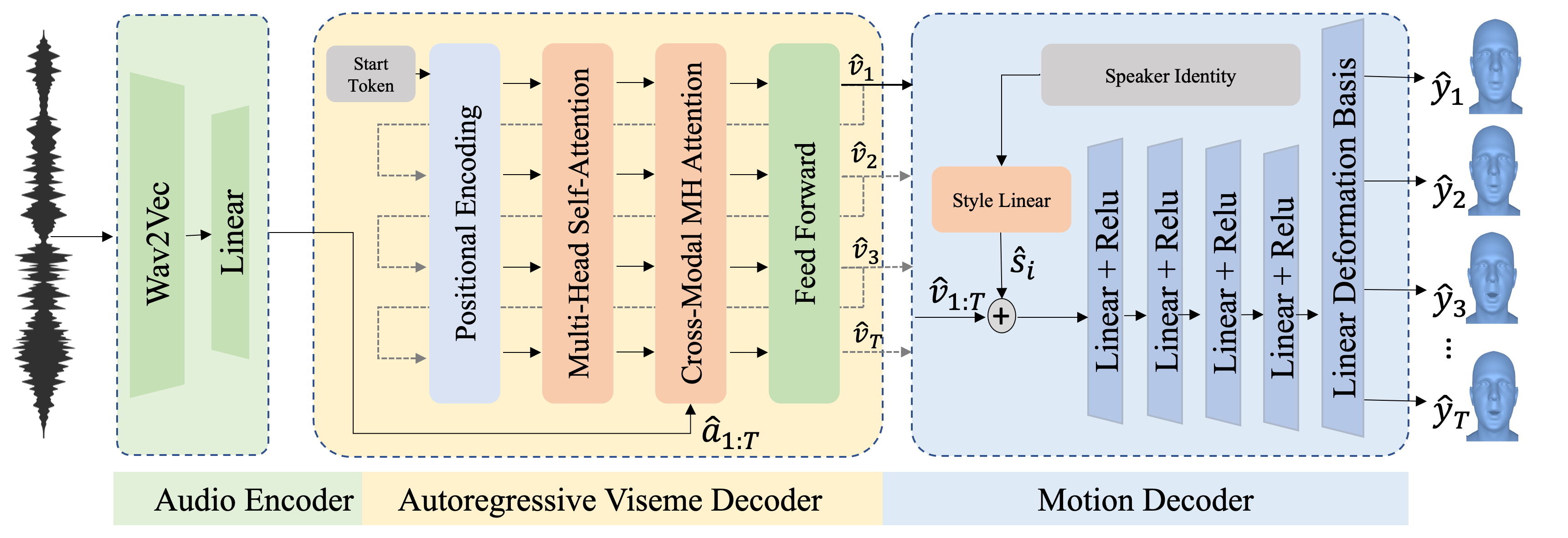}
    \vspace{-0.7cm}
    \caption{
    Our architecture takes audio as input which is encoded by a pre-trained Wav2Vec2.0 model~\cite{wav2vec2.0}.
    This audio embedding $\hat{a}_{1:T}$ is interpreted by an auto-regressive viseme decoder which generates a generalized motion feature $\hat{v}_{1:T}$.
    A style-adaptable motion decoder maps these motion features to person-specific facial expressions $\hat{y }_{1:T}$ in terms of vertex displacements on top of a template mesh.
    }
    \vspace{-0.2cm}
    \label{fig:pipeline}
\end{figure*}

\section{Related Work}
\label{sec:related}

Our work focuses on speech-driven 3D facial animation related to talking head methods that create photo-realistic video sequences from audio inputs.
%

\vspace{-0.1cm}
\paragraph{Talking Head Videos:}
Several prior works on speech-driven generation focus on the synthesis of 2D talking head videos. 
Suwajanakorn et al.~\cite{suwajanakorn2017synthesizing} train an LSTM network on $19$h video material of Obama to predict his person-specific 2D lip landmarks from speech inputs, which is then used for image generation. 
Vougioukas et al.~\cite{vougioukas2020realistic}  propose a method to generate facial animation from a single RGB image by leveraging a temporal generative adversarial network. 
Chung et al.~\cite{chung2017you} introduce a real-time approach to generate an RGB video of a talking face by directly mapping the audio input to the video output space. This method can re-dub a new target identity not seen during training. 
Instead of performing direct mapping, Zhou et al.~\cite{zhou2020makelttalk} disentangles the speech information in terms of speaker identity and content, allowing speech-driven generation that can be applied to various types of realistic and hand-drawn head portraits. 
A series of work~\cite{thies2020nvp,song2022everybody,zhang2021flow,yi2020audio} uses an intermediate 3D Morphable Model (3DMM) \cite{blanz1999morphable, egger20203d} to guide the 2D neural rendering of talking heads from audio.
Wang et al.~\cite{wang2021audio2head} extend this work also to model the head movements of the speaker.
Lipsync3d~\cite{lipsync3d} proposes data-efficient learning of personalized talking heads focusing on pose and lighting normalization.
Based on dynamic neural radiance fields~\cite{nerface}, Ad-nerf~\cite{guo2021adnerf} and DFA-NeRF~\cite{yao2022dfa} learn personalized talking head models that can be rendered under novel views, while being controlled by audio inputs.
In contrast to these methods, our work focuses on predicting 3D facial animations from speech that can be used to drive 3D digital avatars without requiring retraining of the entire model to capture the person-specific motion style.
%

\vspace{-0.1cm}
\paragraph{Speech-Driven 3D Facial Animation:}
Speech-driven 3d facial animation is a vivid field of research.
Earlier methods \cite{de_martino_facial_2006, ezzat_miketalk:_1998, kalberer_face_2001, JALI, verma_using_2003} focus on animating a predefined facial rig using procedural rules.
HMM-based models generate visemes from input text or audio, and the facial animations are generated using viseme-dependent co-articulation models~\cite{JALI, de_martino_facial_2006} or by blending facial templates~\cite{kalberer_face_2001}.
With recent advances in machine learning, data-driven methods~\cite{cao, gen-speech-animation, thies2020nvp, karras_audio-driven_2017, voca, meshtalk, faceformer} have demonstrated their capability to learn viseme patterns from data.
These methods are based on pretrained speech models~\cite{deepspeech,wav2vec,wav2vec2.0} to generate an abstract and generalized representation of the input audio, which is then interpreted by a CNN or auto-regressive model to map to either a 3DMM space or directly to 3D meshes.
Karras et al.\cite{karras_audio-driven_2017} learn a 3D facial animation model from 3-5 minutes of high-quality actor specific 3D data.
VOCA~\cite{voca} is trained on 3D data of multiple subjects and can animate the corresponding set of identities from input audio by providing a one-hot encoding during inference that indicates the subject.
MeshTalk~\cite{meshtalk} is a generalized method that learns a categorical representation for facial expressions and auto-regressively samples from this categorical space to animate a given 3D facial template mesh of a subject from audio inputs.
FaceFormer~\cite{faceformer} uses a pretrained Wav2Vec~\cite{wav2vec2.0} audio representation and applies a transformer-based decoder to regress displacements on top of a template mesh.
Like VOCA, FaceFormer provides a speaker identification code to the decoder, allowing one to choose from the training set talking styles.
In contrast, we aim at a method that can adapt to new users, capturing their talking style and expressiveness.

\section{Method}
\label{sec:main}
Our goal is to model person-specific speaking style and the facial idiosyncrasies of an actor, to generate 3D facial animations of the subject from novel audio inputs.
As input, we assume a short video sequence of the subject which we leverage to compute the identity-specific speaking style. 
To enable fast adaptation to novel users without significant training sequences, we learn a generalized style-agnostic transformer on VOCAset~\cite{voca}.
This transformer provides generic motion features from audio inputs that are interpretable by a person-specific motion decoder.
The motion decoder is pre-trained and adaptable to new users via speaking style optimization and refinement of the motion basis.
To further improve synthesis results, we introduce a novel lip contact loss based on physiological cues of the bilabial consonants~\cite{JALI}.
In the following, we will detail our model architecture and the training objectives and describe the style adaptation.


\subsection{Model Architecture} \label{sec:sty_agn_method}
Our architecture consists of three main components (see \Cref{fig:pipeline}): an audio encoder, a generalized auto-regressive viseme decoder, and an adaptable motion decoder.

\paragraph{Audio Encoder:}
Following state-of-the-art motion synthesis models~\cite{voca, faceformer}, we use a generalized speech model to encode the audio inputs $A$.
Specifically, we leverage the Wav2Vec 2.0 model~\cite{wav2vec2.0}.  
The original Wav2Vec is based on a CNN architecture designed to produce a meaningful latent representation of human speech.
To this end, the model is trained in a self-supervised and semi-supervised manner to predict the immediate future values of the current input speech by using a contrastive loss, allowing the model to learn from a large amount of unlabeled data.
Wav2Vec 2.0 extends this idea by quantizing the latent representation and incorporating a Transformer-based architecture \cite{vaswani2017attention}.
We resample the Wav2Vec 2.0 output with a linear interpolation layer to match the sampling frequency of the motion (30fps for the VOCAset, with 16kHz audio), resulting in a contextual representation $\{\hat{a}\}^{T}_{t=1}$ of the audio sequence for $T$ motion frames.

\paragraph{Auto-regressive Viseme Decoder:} \label{sec:viseme_decoder}
The decoder $F_{v}$ takes the contextual representation of the audio sequence as input and produces style agnostic viseme features $\hat{v}_t$ in an auto-regressive manner.
These viseme features describe how the lip should deform given the context audio and the previous viseme features.
In contrast to Faceformer\cite{faceformer}, we propose to use of a classical transformer architecture~\cite{vaswani2017attention} as viseme decoder, which learns the mapping from audio-features $\{\hat{a}\}^{T}_{t=1}$ to identity agnostic viseme features $\{\hat{v}\}^{T}_{t=1}$.
The autoregressive viseme decoder is defined as:
\begin{equation}
 \hat{v}_t = F_v(\theta_v; \hat{v}_{1:t-1}, \hat{a}_{1:T}),
\end{equation}
\noindent
where $\theta_v$ are the learnable parameters of the transformer.
In contrast to the traditional neural machine translation (NMT) architectures that produce discrete text, our output representation is a continuous vector.
NMT models use a start and end token to indicate the beginning and end of the sequence. During inference, the NMT model auto-regressively generates tokens until the end token is generated. 
Similarly, we use a start token to indicate the beginning of the sequences.
However, since the sequence length $T$ is given by the length of the audio input, we do not use an end token.
We inject temporal information into the sequences by adding encoded time to the viseme feature in the sequence.
We formulate the positionally encoded intermediate representations $\hat{h}_t$ as:
\begin{equation}
 \hat{h}_t = \hat{v}_t + PE(t),
\end{equation}
where $PE(t)$ is a sinusoidal encoding function~\cite{vaswani2017attention}.  
Given the sequence of positional encoded inputs $\hat{h}_t$, we use multi-head self-attention which generates the context representation of the inputs by weighting the inputs based on their relevance.
These context representations are used as input to a cross-modal multi-head attention block which also takes the audio features $\hat{a}_{1:T}$ from the audio encoder as input.
A final feed-forward layer maps the output of this audio-motion attention layer to the viseme embedding $\hat{v}_t$.
In contrast to Faceformer~\cite{faceformer}, which feeds encoded face motions $\hat{y}_t$ to the transformer, we work with identity-agnostic viseme features which are independently decoded by the motion decoder.
We found that feeding face motions $\hat{y}_t$ via an input embedding layer to the transformer contains identity-specific information, which we try to avoid since we aim for a generalized viseme decoder that is disentangled from person-specific motion.
In addition, using a general start token instead of the identity code~\cite{faceformer} as the start token reduces the identity bias further. 
Note that disentangling the identity-specific information from the viseme decoder improves the motion optimization in the style adaption stage of the pipeline (see \Cref{sec:adaptation}), as gradients do not need to be propagated through the auto-regressive transformer.
%


\paragraph{Motion Decoder:}
The motion decoder aims to generate 3D facial animation $\hat{y}_{1:T}$ from the style-agnostic viseme features  $\hat{v}_{1:T}$ and a style embedding $\hat{S}_{i}$.
Specifically, our motion decoder consists of two components, a style embedding layer and a motion synthesis block.
For the training of the style-agnostic transformer and for pre-training the motion decoder, we assume to have a one-hot encoding of the identities of the training set.
The style embedding layer takes this identity information as input and produces the style embedding $\hat{S}_{i}$, which encodes the identity-specific motion.
The style embedding is concatenated with the viseme features $\hat{v}_{1:T}$ and fed into the motion synthesis block.
The motion synthesis block consists of non-linear layers which map the style-aware viseme features to the motion space defined by a linear deformation basis.
During training, the deformation basis is learned across all identities in the dataset.
The deformation basis is fine-tuned for style adaptation to out-of-training identities (see \Cref{sec:adaptation}).
The final mesh outputs $\hat{y}_{1:T}$ are computed by adding the estimated per-vertex deformation to the template mesh of the subject.
%


\subsection{Training}
Similar to Faceformer~\cite{faceformer}, we use an autoregressive training scheme instead of teacher-forcing to train our model on the VOCAset~\cite{voca}.
Given that VOCAset provides ground truth 3D facial animations, we define the following loss:
\begin{equation}
    \label{eq:total}
 \mathcal{L}_{total} = \lambda_{MSE}\cdot\mathcal{L}_{MSE}  + \lambda_{vel}\cdot\mathcal{L}_{vel} + \lambda_{lip}\cdot\mathcal{L}_{lip},
\end{equation}
where $\mathcal{L}_{MSE}$ defines a reconstruction loss of the vertices, $\mathcal{L}_{vel}$ defines a velocity loss, and $\mathcal{L}_{lip}$ measures lip contact.
The weights are $\lambda_{MSE}=1.0$, $\lambda_{vel}=10.0$, and $\lambda_{lip}=5.0$.

\paragraph{Reconstruction Loss:}
The reconstruction loss $\mathcal{L}_{MSE}$ is:
\begin{equation}
 \mathcal{L}_{MSE} = \sum_{v=1}^{V} \sum_{t=1}^{T_v} || y_{t,v} - \hat{y}_{t,v} || ^2,
\end{equation}
where $y_{t,v}$ is the ground truth mesh at time $t$ in sequence $v$ (of $V$ total sequences) and $\hat{y}_{t,v}$ is the prediction.

\paragraph{Velocity Loss:}
Our motion decoder takes independent viseme features as input to produce facial expressions.
To improve temporal consistency in the prediction, we introduce a velocity loss $\mathcal{L}_{vel}$ similar to \cite{voca}:
\begin{equation}
 \mathcal{L}_{vel} = \sum_{v=1}^{V} \sum_{t=2}^{T_v}  || (y_{t,v} - y_{t-1,v}) - (\hat{y}_{t,v} - \hat{y}_{t-1,v}) || ^2.
\end{equation}
\noindent

\paragraph{Lip Contact Loss:}
Training with $L_{MSE}$ guides the model to learn an averaged facial expression, thus resulting in improper lip closures.
To this end, we introduce a novel lip contact loss for bilabial consonants ('m','b','p') to improve lip closures.
Specifically, we automatically annotate the VOCAset to extract the occurrences of these consonants; see \Cref{sec:data}.
Using this data, we define the following lip loss:
\begin{equation}
 \mathcal{L}_{lip} = \sum_{t=1}^{T} \sum_{j=1}^{N} w_t || y_{t,v} - \hat{y}_{t,v} || ^2,
\end{equation}
where $w_{t,v}$ weights the prediction of frame $t$ according to the annotation of the bilabial consonants.
Specifically, $w_{t,v}$ is one for frames with such consonants and zero otherwise.
Note that for such consonant frames, the target $y_{t,v}$ represents a face with a closed mouth; thus, this loss improves lip closures at 'm','b' and 'p's (see \Cref{sec:results}).
%

\subsection{Style Adaptation}
\label{sec:adaptation}

Given a video of a new subject, we reconstruct and track the face $\Tilde{y}_{1:T}$ (see \Cref{sec:video_processing}).
Based on this reference data, we first optimize for the speaker style-embedding $\hat{S}$ and then jointly refine the linear deformation basis using the $\mathcal{L}_{MSE}$ and $\mathcal{L}_{vel}$ loss.
In our experiments, we found that this two-stage adaptation is essential for generalization to new audio inputs as it reuses the pretrained information of the motion decoder.
As an initialization of the style embedding, we use a speaking style of the training set.
We precompute all viseme features $\hat{v}_{1:T}$ once, and optimize the speaking style to reproduce the tracked faces $\Tilde{y}_{1:T}$.
We then refine the linear motion basis of the decoder to match the person-specific deformations (e.g., asymmetric lip motions).
%

\section{Dataset}
\label{sec:data}
\label{sec:mbp_detection}
\label{sec:video_processing}

We train our method based on the VOCAset~\cite{voca}, which consists of $12$ actors ($6$ female and $6$ male) with $40$ sequences each with a length of $3-5$ seconds.
The dataset comes with a train/test set split which we use in our experiments.
The test set contains $2$ actors.
The dataset offers audio and high-quality 3D face reconstructions per frame ($60$fps). For our experiment, we sample the 3D face reconstructions at $30$fps.
We train the auto-regressive transformer on this data using the loss from \Cref{eq:total}.
For the lip contact loss $L_{lip}$, we automatically compute the labels as described below.
To adapt the motion decoder to a new subject, we require a short video clip of the person.
Using this sequence, we run a 3DMM-based face tracker to get the per-frame 3D shape of the person.
Based on this data, we adapt the motion decoder as detailed in \Cref{sec:adaptation}.

\paragraph{Automatic Lip Closure Labeling:} 
\begin{figure}[t!]
    \centering
    \includegraphics[width=\linewidth]{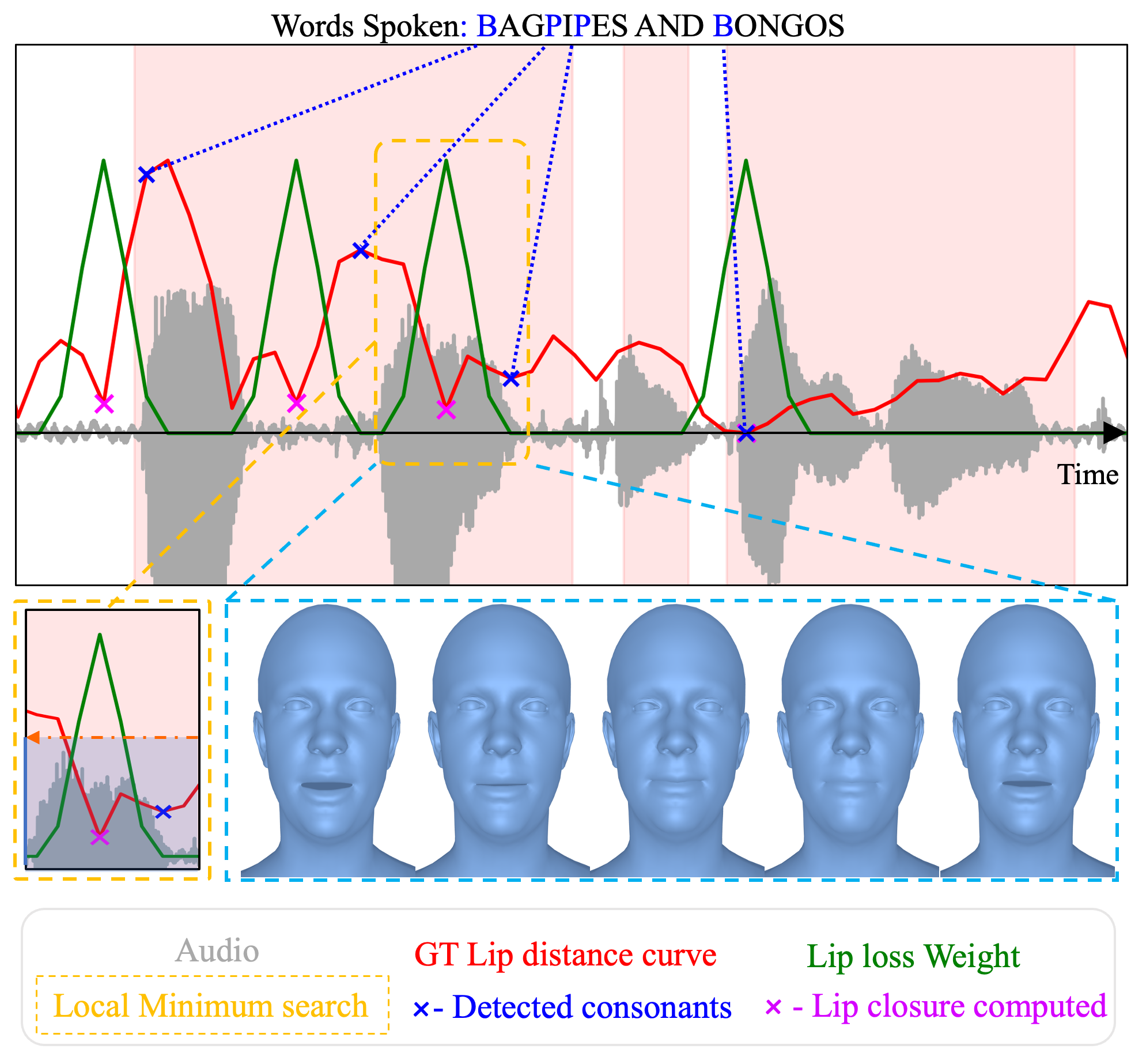}
    \vspace{-0.75cm}
      \caption{
      Automatic labeling of the bilabial consonants ('m','b' and 'p') and their corresponding lip closures in a sequence of VOCAset~\cite{voca}. 
      We align the transcript with the audio track using  Wav2vec~\cite{wav2vec2.0} features and extract the time stamps for the bilabial consonants.
      To detect the lip closures for the bilabial consonants, we search for local-minima on the Lip distance curves (\textcolor{red}{red}).
      The lip loss weights \textcolor{green}{$w_{t,v}$} in a window around the detected lip closure are set to fixed values of a Gaussian function.
      We show an example of detected lip closures in the figure (in the blue bounding box).  
      }
      \vspace{-0.25cm}
      \label{fig:automatic_labeling}
\end{figure}
For the VOCAset, the transcript is available.
Based on Wav2Vec features, we align the transcript with the audio track.
As the lip closure is formed before we hear the bilabial consonants, we search for the lip closure in the tracked face geometry before the time-stamp of the occurrence of the consonants in the script.
We show this process for a single sequence in \Cref{fig:automatic_labeling}.
The lip closure is detected by lip distance, i.e., the frame with minimal lip distance in a short time window before the consonant is assumed to be the lip closure.

\paragraph{External Sequence Processing:} 
We assume to have a monocular RGB video of about $2$ minutes in length as input which we divide into train/validation/test sequences.
Based on MICA~\cite{mica}, we estimate the 3D shape of the subject using the first frame of the video.
Using this shape estimate, we run an analysis-by-synthesis approach~\cite{face2face} to estimate per-frame blendshape parameters of the FLAME 3DMM~\cite{flame}.
Given these blendshape coefficients, we can compute the 3D vertices of the per-frame face meshes that we need to adapt the motion decoder.
Note that in contrast to the training data of the transformer, we do not require any bilabial consonants labeling, as we adapt the motion decoder only based on the reconstruction and velocity loss.

\begin{figure*}[ht!]
    \includegraphics[width=\textwidth]{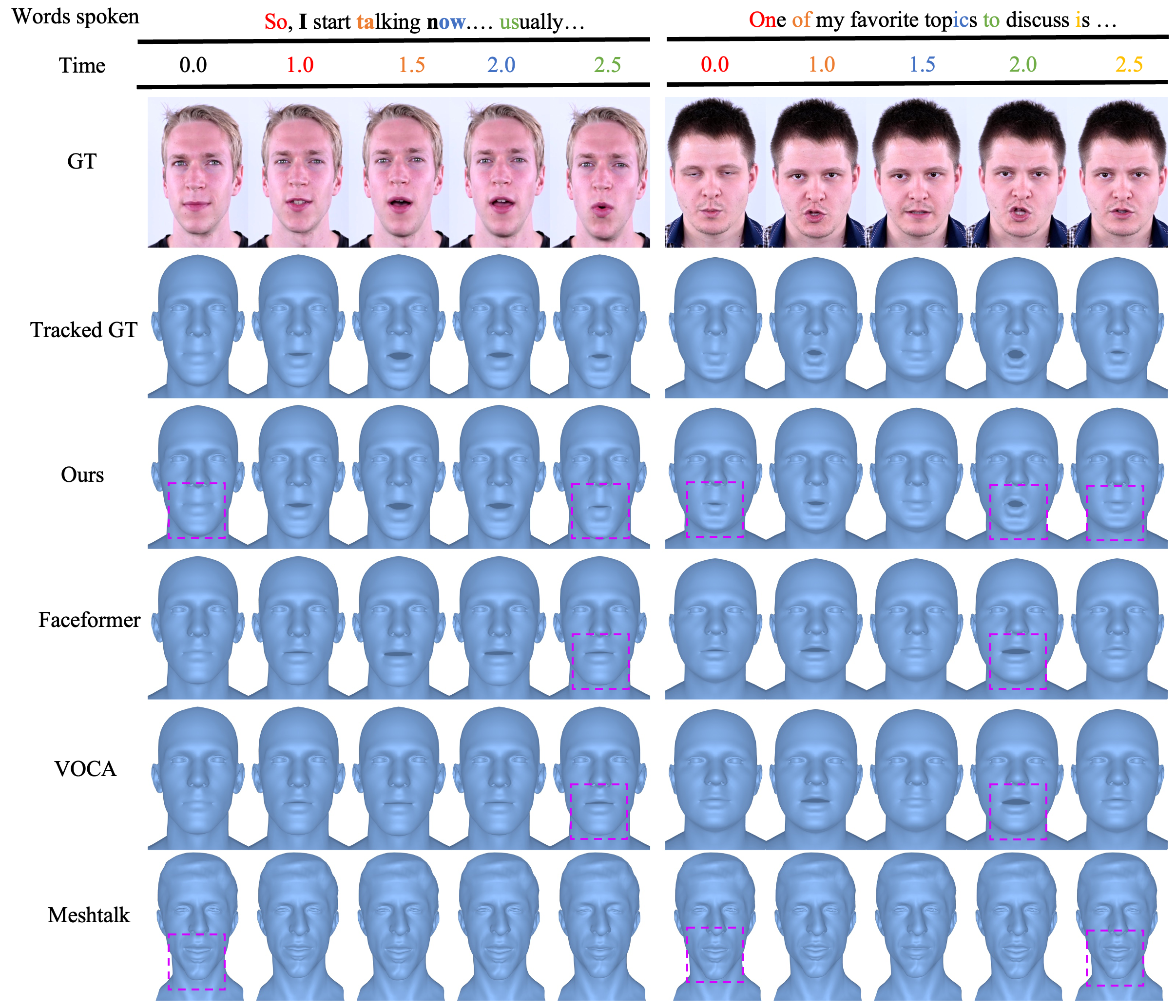}
     \vspace{-0.5cm}
      \caption{
      Qualitative comparison to the state-of-the-art methods VOCA~\cite{voca}, Faceformer~\cite{faceformer}, and MeshTalk~\cite{meshtalk}.
      Note that \textit{MeshTalk is performed with a different identity} since we use their pretrained model, which cannot be trained on VOCAset.
      As we see in the highlighted regions, the geometry of the generated sequences without the person-specific style have muted and inaccurate lip animations. 
      }
      \vspace{-0.25cm}
      \label{fig:qualitative}
\end{figure*}

\section{Results}
\label{sec:results}

\begin{figure*}[ht!]
    \centering
    \vspace{-0.3cm}
    \includegraphics[width=0.9\textwidth]{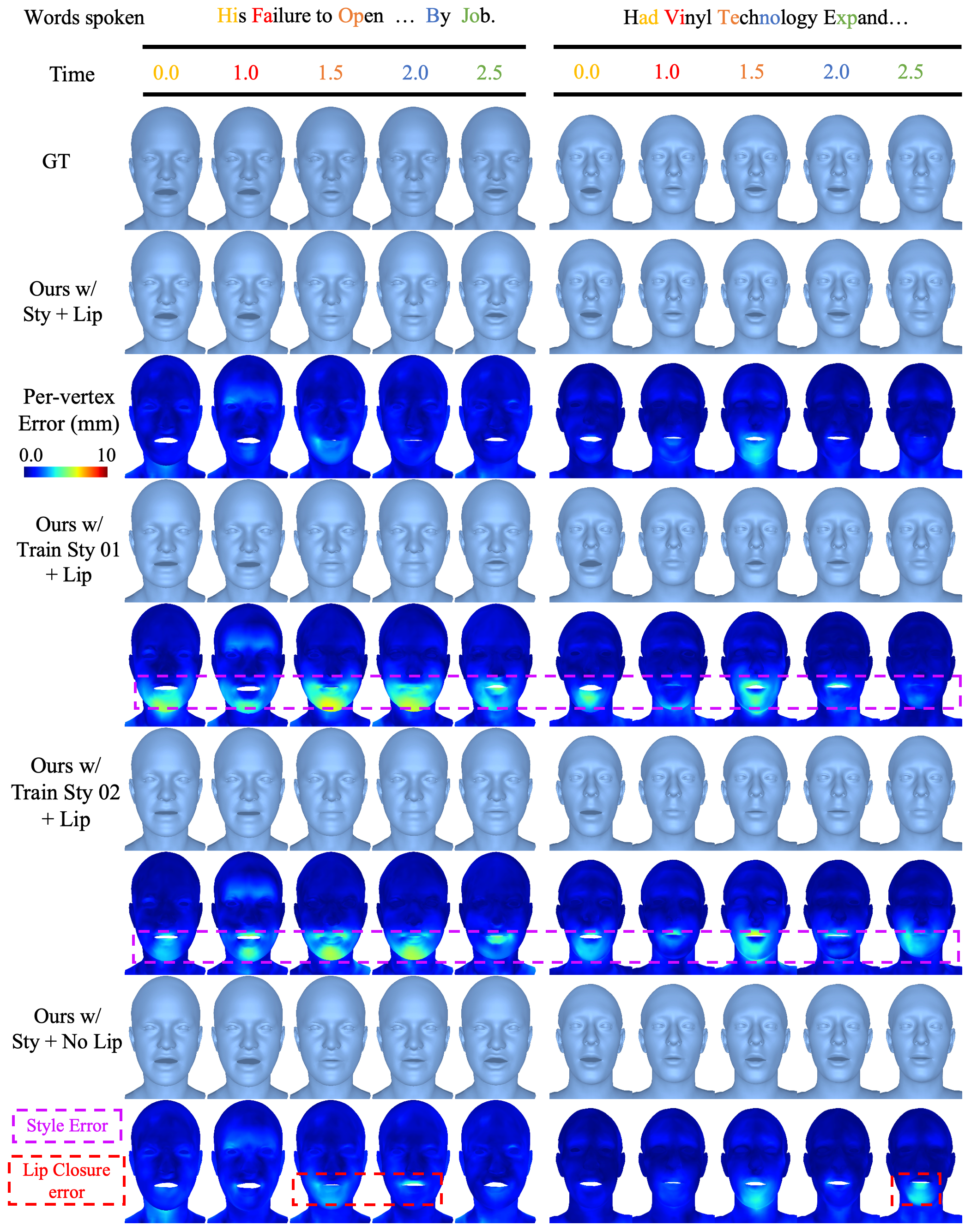}
      \caption{
      Qualitative ablation comparison.
      At first, we show that our complete method with style and $\mathcal{L}_{lip}$ loss is able to generate personalized facial animation with expressive motion and accurate lip closures.
      Replacing the person-specific style with the style seen during training results in generic and muted facial animation.
      As highlighted in the per-vertex error maps (\textcolor{magenta}{magenta}), the generated expression is not similar to the target actor.
      Especially the facial deformations are missing person-specific details.
      Removing $\mathcal{L}_{lip}$ from the training objective results in improper lip closures (\textcolor{red}{red}).
      }
      \label{fig:ablation_main}
\end{figure*}

To validate our method, we conducted a series of qualitative and quantitative evaluations, including a user study and ablation studies.
For evaluation on the test set of VOCAset~\cite{voca}, we randomly sample $4$ sequences from the test subjects' train set (each $\sim5$s long) and learn the speaking-style and facial idiosyncrasies of the subject via style adaptation.
We compare our method to the state-of-the-art methods VOCA~\cite{voca}, Faceformer~\cite{faceformer}, and MeshTalk~\cite{meshtalk}.
We use the original implementations of the authors. However, we found that MeshTalk cannot train on the comparably small VOCAset.
Thus, we qualitatively compare against MeshTalk with their provided model trained on a large-scale proprietary dataset with $200$ subjects and $40$ sequences for each.
Note that the pretrained MeshTalk model is not compatible with the FLAME topology; thus, we cannot evaluate their method on novel identities.
In addition to the experiments on the VOCAset, we show results on external RGB sequences.
The results can be best seen in the suppl. video.

\begin{table}[b]
    \vspace{-.5cm}
    \resizebox{\linewidth}{!}{%
        \begin{tabular}{l|ccccc} \toprule
            \textbf{Method}& \textbf{$L_2^{face}$} $\downarrow$& \textbf{$L_2^{lip}$} $\downarrow$& \textbf{F-DTW} $\downarrow$& \textbf{Lip-DTW} $\downarrow$& \textbf{Lip-sync} $\downarrow$ \\ \midrule
            VOCA\cite{voca} & $0.88$ &  $0.15$ &  $1.28$ & $2.41$ & $5.72$ \\
            Faceformer\cite{faceformer} & $\mathbf{0.8}$ &  $0.14$ &  $\mathbf{1.18}$  & $2.85$ & $5.41$ \\
            Ours (w/ 1seq) & $0.91$ &  $0.1$ &  $1.3$ & $1.68$ & $3.99$ \\
            Ours  & $0.89$ &  $\mathbf{0.09}$ &  $1.26$ &  $\mathbf{1.47}$ & $\mathbf{3.78}$ \\
            \bottomrule
        \end{tabular}
    }%
    \vspace{-0.1cm}
    \caption{Quantitative results on the VOCAset~\cite{voca}. Our method outperforms the baselines on all of the lip metrics while performing on par on the full-face metrics.
    Note that we are not targeting the animation of the upper face but aim for expressive and accurate lip movements, which is noticeable from the improved lip scores.
    } 
    \label{tab:qunatitative_study}
\end{table}

%
\paragraph{Quantitative Evaluation:}
To quantitatively evaluate our method, we use the test set of VOCAset~\cite{voca}, which provides high-quality reference mesh reconstructions.
We evaluate the performance of our method based on a mean $L_2$ vertex distance for the entire mesh $L_2^{face}$ and the lip region $L_2^{lip}$.
Following MeshTalk~\cite{meshtalk}, we also compute the Lip-sync, which measures the mean of the maximal per-frame lip distances.
In addition, we use Dynamic Time Wrapping (DTW) to compute the similarity between the produced and reference meshes, both for the entire mesh (F-DTW) and the lip region (Lip-DTW).
Since VOCA and Faceformer do not adapt to new user talking styles, we select the talking style from their training with the best quantitative metrics.
Note that the pretrained MeshTalk model is not applicable to this evaluation due to the identity mismatch.
As can be seen in \Cref{tab:qunatitative_study}, our method achieves the lowest lip reconstruction and lip-sync errors, confirming our qualitative results.
Even when using a single reference video for style adaptation (5s), our results shows significantly better lip scores.

\paragraph{Qualitative Evaluation:}
We conducted a qualitative evaluation on external sequences not part of VOCAset.
In \Cref{fig:qualitative}, we show a series of frames from those sequences with the corresponding words.
As we can see, our method is able to adapt to the speaking style of the respective subject.
VOCA~\cite{voca} and Faceformer~\cite{faceformer} miss person-specific deformations and are not as expressive as our results.
MeshTalk~\cite{meshtalk}, which uses an identity that comes with the pretrained model, also shows dampened expressivity.
In the suppl. video, we can observe that our method is generating better lip closures for bilabial consonants.

\paragraph{Perceptual Evaluation:}
We conducted a perceptual evaluation to quantify the quality of our method's generated results (see \Cref{tab:perceptual_study}).
Specifically, we conducted an A/B user study on the test set of VOCAset.
We randomly sample 10 sequences of the test subjects and run our method, VOCA, and Faceformer.
For VOCA and Faceformer, which do not adapt to the style of a new user, we use the talking style of the training Subject 137, which provided the best quantitative results.
We use 20 videos per method resulting in 60 A/B comparisons.
For every A/B test, we ask the user to choose the best method based on realism and expressiveness, following the user study protocol of Faceformer~\cite{faceformer}.
In \Cref{tab:perceptual_study}, we show the result of this study in which $56$ people participated.
We observe that our method consistently outperforms VOCA and Faceformer.
We also see that our model achieves similar realism and lip-sync as ground truth.
Note that the users in the perceptual study have not seen the original talking style of the actors before. However, the results show that our personalized synthesis leads to more realistic-looking animations.
 \begin{table}
    \resizebox{\linewidth}{!}{%
        \begin{tabular}{l|cccc} \toprule
            \textbf{Method}& \textbf{Expressiveness} (\%) &  \textbf{Realism/Lip-sync} (\%) \\ \midrule
            Ours vs VOCA~\cite{voca} & $86.48$& $76.92$ \\
            Ours vs Faceformer~\cite{faceformer} & $81.89$ & $75.46$ \\
            Ours vs Ground truth  & $20.28$ & $42.30$ \\
            \bottomrule
        \end{tabular}
    }
    \caption{
    In a perceptual A/B user study conducted on the test set of VOCAset~\cite{voca} with $56$ participants, we see that in comparison to VOCA~\cite{voca} and Faceformer~\cite{faceformer} our method is preferred.
    }
    \label{tab:perceptual_study}
\end{table}

\subsection{Ablation Studies}

To understand the impact of our style adaptation and the novel lip contact loss $\mathcal{L}_{lip}$ on the perceptual quality, we show a qualitative ablation study including per-vertex error maps in \Cref{fig:ablation_main}.
As highlighted in the figure, the style adaptation is critical to match the person-specific deformations and mouth shapes and improves expressiveness.
The lip contact loss improves the lip closures for the bilabial consonants, thus, improving the perceived realism, as can best be seen in the suppl. video.
We rely on only $\sim60$ seconds-long reference videos to extract the person-specific speaking style.
A detailed analysis of the sequence length's influence on the final output quality can be found in the suppl. material. 
It is also worth noting that our style-agnostic architecture allows us to perform style adaptation of the motion decoder in less than $30$min, while an adaptation with an identity-dependent transformer takes about $6$h.

\begin{figure}[t!]
    \centering
    \includegraphics[width=\linewidth]{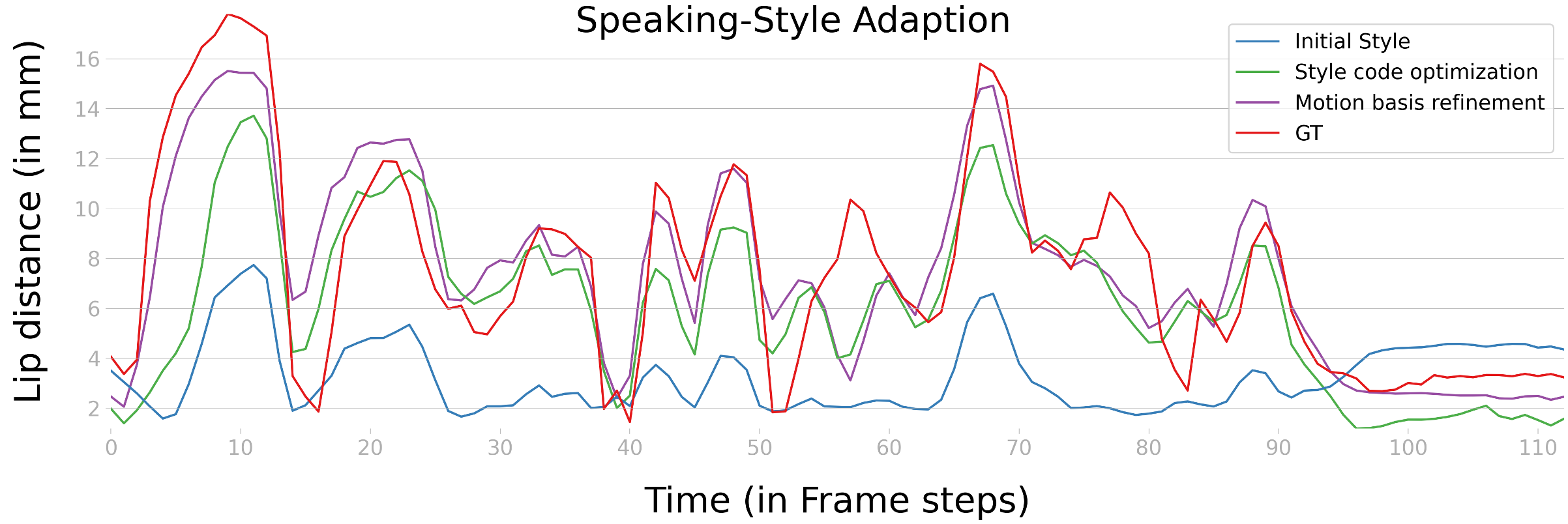}
      \caption{
      Analysis of style adaptation in terms of lip distance on a test sequence of the VOCAset~\cite{voca} (reference in \textcolor{red}{red}).
      Starting from an initial talking style from the training set (\textcolor{blue}{blue}), we consecutively adapt the style code (\textcolor{green}{green}) and the motion basis of the motion decoder (\textcolor{purple}{purple}).
      }
      \label{fig:2_stage_style_adap}
\end{figure}

Our proposed style adaptation has two stages as explained in \Cref{sec:adaptation}.
In the first step, we optimize for the style code and the refine the motion basis.
In \Cref{fig:2_stage_style_adap}, we show an example of the style adaptation by evaluating the lip distances throughout a sequence with a motion decoder at initialization, with optimized style code, and with a refined motion basis.
While the lip distance with the generalized motion decoder is considerable, it gets significantly improved by the consecutive steps of style adaptation.
After style code optimization, we observe that the amplitude and frequency of the lip distance curves start resembling the ground truth.
Refining the motion basis further improves the lip distance, and it is able to capture facial idiosyncrasies, like asymmetrical lip deformations.
\section{Discussion}
\label{sec:discussion}

Our evaluation shows that our proposed method outperforms state-of-the-art methods in perceived expressiveness and realism.
However, several limitations remain. 
Specifically, we only support the speaking style of the subject seen in the reference video and do not control the talking style w.r.t. emotions (e.g., sad, happy, angry).
The viseme transformer and the motion decoder could be conditioned on an emotion flag; we leave this for future work.
The expressiveness and facial details depend on the face tracker's quality; if the face tracking is improved, our method will predict better face shapes. 

\section{Conclusion}
\label{sec:conclusion}

We present \textit{Imitator}, a novel approach for personalized speech-driven 3D facial animation.
Based on a short reference video clip of a subject, we learn a personalized motion decoder driven by a generalized auto-regressive transformer that maps audio to intermediate viseme features.
Our studies show that personalized facial animations are essential for the perceived realism of a generated sequence.
Our new loss formulation for accurate lip closures of bilabial consonants further improves the results.
We believe that personalized facial animations are a stepping stone towards audio-driven digital doubles.
\section{Acknowledgements}
This project has received funding from the Mesh Labs, Microsoft, Cambridge, UK.
Further, we would like to thank Berna Kabadayi, Jalees Nehvi, Malte Prinzler and Wojciech Zielonka for their support and valuable feedback. 
The authors thank the International Max Planck Research School for Intelligent Systems (IMPRS-IS) for supporting Balamurugan Thambiraja.

{\small
\bibliographystyle{splncs04}
\bibliography{main}
}

\clearpage
\twocolumn[{%
\renewcommand\twocolumn[1][]{#1}%
\begin{center}
     \textbf{\Large{\methodtitle \\ -- Supplemental Document --}}
     \vspace{1cm}
\end{center}%
}]


\section{Impact of Data to Style Adaptation:}
\label{sec:style_adapt}

To analyze the impact of data on the style adaptation process, we randomly sample ($1, 4, 10, 20 $) sequences from the train set of the VOCA test subjects and perform our style adaption.
Each sequence contains about $3-5$ seconds of data.
In \Cref{tab:seq_ablation_study}, we observe that the performance on the quantitative metrics increase with the number of reference sequences.
As mentioned in the main paper, even an adaptation based on a single sequence results in a significantly better animation in comparison to the baseline methods.
This highlights the impact of style on the generated animations.

\begin{figure}[ht!]
    \centering
    \includegraphics[width=\linewidth]{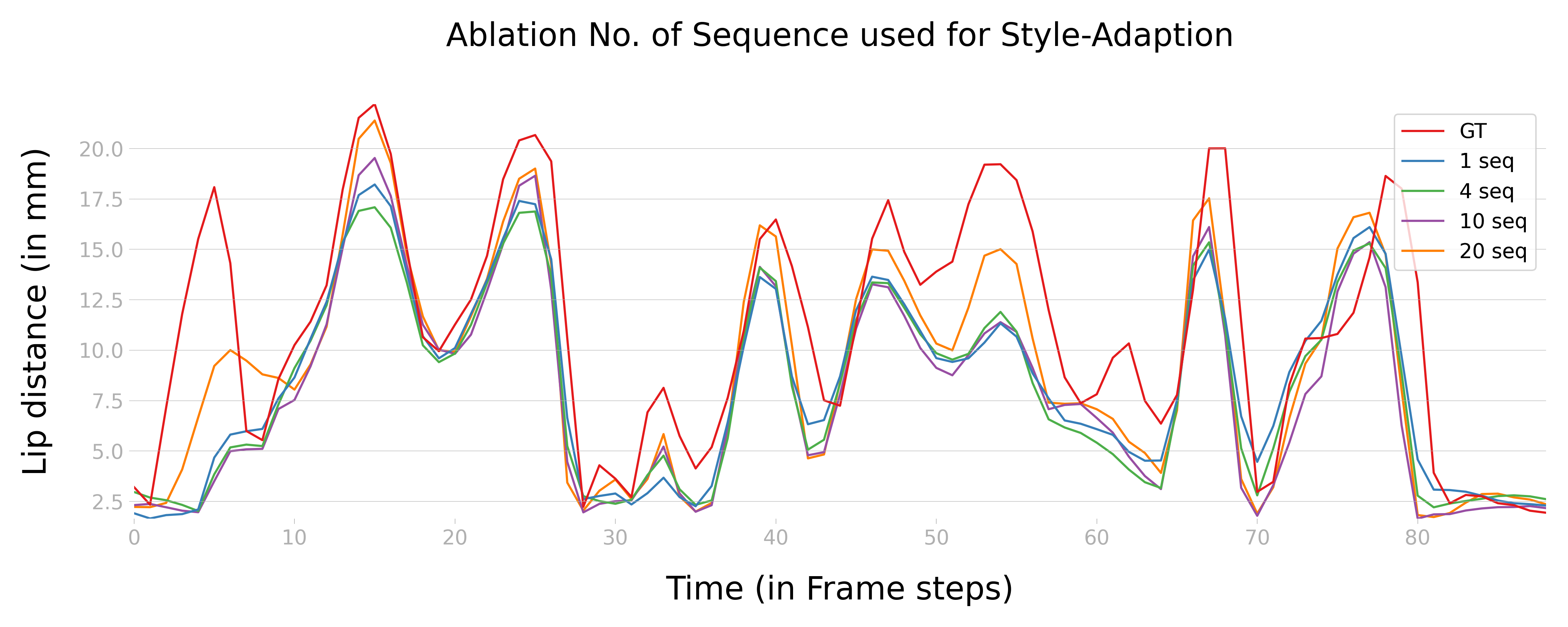}
      \caption{
      With an increasing number of reference data samples for style adaptation, the lip distance throughout a test sequence of VOCAset is approaching the ground truth lip distance curve.
      }
      \vspace{-0.25cm}
      \label{fig:no_seq_style_adap}
\end{figure}

\Cref{fig:no_seq_style_adap} illustrates the lip distance curve for one test sequence used in this study.
We observe that the lip distance with more reference data better fits the ground truth curve.

 \begin{table}[h]
    \resizebox{\linewidth}{!}{%
    \begin{tabular}{l|ccccc} \toprule
            \textbf{No. Seq.}& \textbf{$L_2^{face}$} $\downarrow$& \textbf{$L_2^{lip}$} $\downarrow$& \textbf{F-DTW} $\downarrow$& \textbf{Lip-DTW} $\downarrow$& \textbf{Lip-sync} $\downarrow$ \\ \midrule
        1 & $0.91$ &  $0.1$ &  $1.3$ & $1.68$ & $3.99$ \\
        4 & $0.89$ &  $0.1$ &  $1.26$ & $1.47$ & $3.78$ \\
        10 &  $0.76$ &  $0.09$ &  $1.07$ & $1.37$ & $3.57$  \\
        20 & $0.7$ &  $0.09$ &  $0.99$ & $1.27$ & $3.49$ \\
        \bottomrule
    \end{tabular}
    }
    \caption{
    Ablation of the style adaptation w.r.t. the amount of reference sequences used.
    With an increasing number of data, the quantitative metrics improve. Each sequence is $3-5$s long.
    }
    \vspace{-0.25cm}
    \label{tab:seq_ablation_study}
\end{table}

\section{Architecture Details}
\label{sec:arch}

\subsection{Audio Encoder:}
Similar to Faceformer\cite{faceformer}, our audio encoder is built upon the Wav2Vec 2.0 \cite{wav2vec2.0} architecture to extract temporal audio features. 
These audio features are fed into a linear interpolation layer to convert the audio frequency to the motion frequency.
The interpolated outputs are then fed into 12 identical transformer encoder layers with 12 attention heads and an output dimension of 768.
A final linear projection layer converts the audio features from the 768-dimension features to a 64-dimensional phoneme representation.

\subsection{Auto-regressive Viseme Decoder:}
Our auto-regressive viseme decoder is built on top of traditional transformer decoder layers ~\cite{vaswani2017attention}.
We use a zero vector of 64-dimension as a start token to indicate the start of sequence synthesis.
We first add a positional encoding of 64-dimension to the input feature and fed it to decoder layers in the viseme decoder.
For self-attention and cross-modal multi-head attention, we use 4 heads of dimension 64.
Our feed forward layer dimension is 128. 

\paragraph{Multi-Head Self-Attention:}
Given a sequence of positional encoded inputs $\hat{h}_t$, we use multi-head self-attention (self-MHA), which generates the context representation of the inputs by weighting the inputs based on their relevance.
The Scaled Dot-Product attention function can be defined as mapping a query and a set of key-value pairs to an output, where queries, keys, values and outputs are vectors \cite{vaswani2017attention}.
The output is the weighted sum of the values; the weight is computed by a compatibility function of a query with the corresponding key.
The attention can be formulated as:
\begin{equation}\label{eqn:attention}
    Attention(Q,K,V) = \sigma(\frac{QK^T}{\sqrt{d_k}})V,
\end{equation}
where $Q, K, V$ are the learned Queries, Keys and Values, $\sigma(\cdot)$ denotes the softmax activation function, and $d_k$ is the dimension of the keys. 
Instead of using a single attention mechanism and generating one context representation, MHA uses multiple self-attention heads to jointly generate multiple context representations and attend to the information in the different context representations at different positions.
MHA is formulated as follows:
\begin{equation}
    MHA(Q,K,V) = [head_1,....,head_h] \cdot W^O,
\end{equation}
with $head_i = Attention(QW^Q_i,KW^K_i,VW^V_i)$, where $W^O,W^Q_i,W^K_i,W^V_i$ are weights related to each input variable.

\paragraph{Audio-Motion Multi-Head Attention}
The Audio-Motion Multi-Head attention aims to map the context representations from the audio encoder to the viseme representations by learning the alignment between the audio and style-agnostic viseme features.
The decoder queries all the existing viseme features with the encoded audio features, which carry both the positional information and the contextual information, thus, resulting in audio context-injected viseme features.
Similar to Faceformer~\cite{faceformer}, we add an alignment bias along the diagonal to the query-key attention score to add more weight to the current time audio features.
The alignment bias $B^{A} (1 \leq i \leq t, 1 \leq j \leq KT )$ is:
\begin{align}\label{eqn:alignbias}
        B^{A}(i, j) =
    \begin{cases}
           0 &\text{if } (i=j), \\
        \hfill -\infty \hfill &\text{otherwise} .
    \end{cases}
\end{align}
The modified Audio-Motion Attention is represented as:
\begin{equation}\label{eqn:audio_motion_attention}
Attention(Q^v,K^a,V^a, B^{A}) = \sigma(\frac{Q^v (K^a)^T}{\sqrt{d_k}} + B^{A})V^a,
\end{equation}
where $Q^v$ are the learned queries from viseme features, $K^a$ the keys and $V^a$ the values from the audio features, $\sigma(\cdot)$ is the softmax activation function, and $d_k$ is the dimension of the keys.

\subsection{Motion Decoder:}
The motion decoder aims to generate 3D facial animations $\hat{y}_{1:T}$ from the style-agnostic viseme features  $\hat{v}_{1:T}$ and a style embedding $\hat{S}_{i}$.
Specifically, our motion decoder consists of two components, a style embedding layer and a motion synthesis block.
The style linear layer takes a one-hot encoder of 8-dimension and produce a style-embedding of 64-dimension.
The input viseme features are concatenated with the style-embedding and fed into 4 successive linear layers which have a leaky-ReLU as activation.
The output dimension of the 4-layer block is 64 dimensional. 
A final fully connected layer maps the 64-dimension input features to the 3D face deformation described as per-vertex displacements of size 15069.
This layer is defining the motion deformation basis of a subject and is adapted based on a reference sequence.

\paragraph{Training Details:}
We use the ADAM optimizer with a learning rate of 1e-4 for both the style-agnostic transformer training and the style adaptation stage.
During the style-agnostic transformer training, the parameters of the Wave2Vec 2.0 layers in the audio encoder are fixed.
Our model is trained for 300 epochs, and the best model is chosen based on the validation reconstruction loss.
During the style-adaptation stage, we first generate the viseme features and keep them fixed during the style adaptation stage. 
Then, we optimize for the style embedding for 300 epochs. Finally, the style-embedding and final motion deformation basis is refined for another 300 epochs.
%


\section{Broader Impact}
\label{sec:broader_impact}
Our proposed method aims at the synthesis of realistic-looking 3D facial animations.
Ultimately, these animations can be used to drive photo-realistic digital doubles of people in audio-driven immersive telepresence applications in AR or VR.
However, this technology can also be misused for so-called DeepFakes.
Given a voice cloning approach, our method could generate 3D facial animations that drive an image synthesis method.
This can lead to identity theft, cyber mobbing, or other harmful criminal acts.
We believe that conducting research openly and transparently could raise the awareness of the misuse of such technology.
We will share our implementation to enable research on digital multi-media forensics.
Specifically, synthesis methods are needed to produce the training data for forgery detection~\cite{roessler2019faceforensics++}.
All participants in the study have given written consent to the usage of their video material for this publication.

\end{document}